\documentclass[11pt]{article}

\usepackage[preprint]{acl}

\usepackage{times}
\usepackage{latexsym}

\usepackage[T1]{fontenc}

\usepackage[utf8]{inputenc}

\usepackage{microtype}

\usepackage{inconsolata}

\usepackage{graphicx}

\usepackage{amssymb}
\usepackage[dvipsnames]{xcolor}
\usepackage{paralist}
\usepackage{amsmath}
\usepackage{booktabs}
\usepackage{listings}
\usepackage{enumitem}
\lstdefinestyle{prompt}{
    basicstyle=\ttfamily\small,
    breaklines=true,
    breakatwhitespace=true,
    breakindent=0pt,
    frame=single,
    framesep=8pt,
    xleftmargin=12pt,
    xrightmargin=12pt,
    backgroundcolor=\color{gray!8},
    rulecolor=\color{gray!40},
    aboveskip=1em,
    belowskip=1em,
    columns=fullflexible,
    keepspaces=true,
}
\makeatletter
\DeclareRobustCommand{\iscircle}{\mathord{\mathpalette\is@circle\relax}}
\newcommand\is@circle[2]{%
  \begingroup
  \sbox\z@{\raisebox{\depth}{$\m@th#1\bigcirc$}}%
  \sbox\tw@{$#1\square$}%
  \resizebox{!}{\ht\tw@}{\usebox{\z@}}%
  \endgroup
}
\makeatother

\title{The Aftermath of DrawEduMath: Vision Language Models \\ Underperform with Struggling Students and Misdiagnose Errors}

\author{
Li Lucy$^{\triangle}$ \quad
\textbf{Albert Zhang}$^+$ \quad
\textbf{Nathan Anderson}$^\div$ \quad
\textbf{Ryan Knight}$^+$ \quad
\textbf{Kyle Lo}$^{\square\triangle}$ \vspace{0.3em}
\\
$^\triangle$University of Washington \quad
$^+$Insource Services \quad
$^\div$Worcester Polytechnic Institute \vspace{0.2em} \\
$^{\square}$Allen Institute for AI 
\vspace{0.3em} \\
\texttt{lucy3li@cs.washington.edu} \quad \texttt{kylel@allenai.org}
}

\begin{document}
\maketitle
\begin{abstract}
Effective mathematics education requires identifying and responding to students' mistakes. For AI to support pedagogical applications, models must perform well across different levels of student proficiency. Our work provides an extensive, year-long snapshot of how 11 vision-language models (VLMs) perform on DrawEduMath, a QA benchmark involving real students' handwritten, hand-drawn responses to math problems. We find that models' weaknesses concentrate on a core component of math education: student error. All evaluated VLMs underperform when describing work from students who require more pedagogical help, and across all QA, they struggle the most on questions related to assessing student error. Thus, while VLMs may be optimized to be math problem solving experts, our results suggest that they require alternative development incentives to adequately support educational use cases. 
\end{abstract}

\section{Introduction}

\begin{figure*}[t]
    \includegraphics[width=\textwidth]{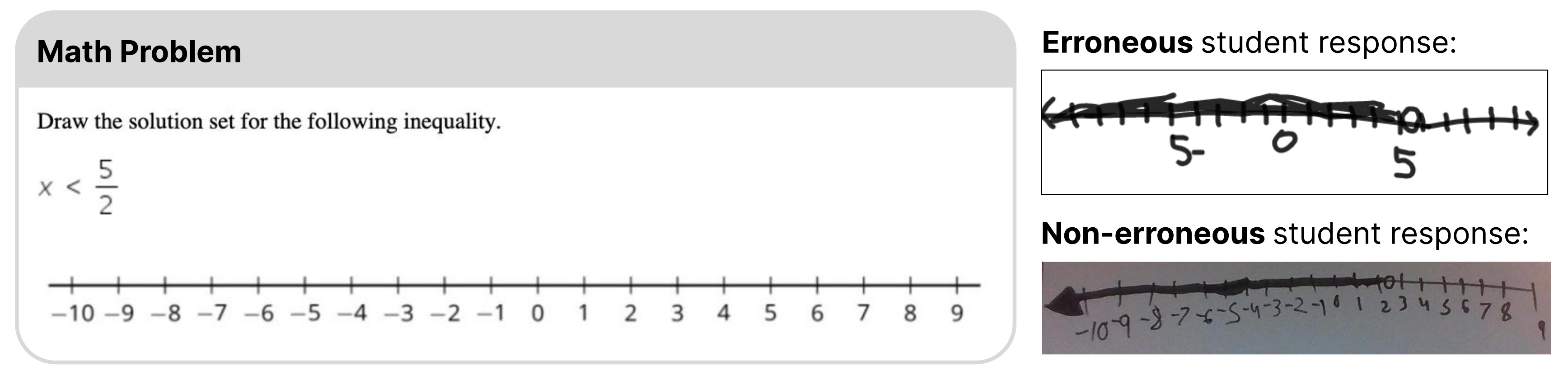}
    \centering
    \caption{On the left is a math problem, where students are asked to draw $x < 5/2$ on a number line. The right side shows two example student responses that differ in correctness. DrawEduMath pairs each math problem with one student response, and prompts VLMs to answer questions about the student response.}
	\label{fig:benchmark_example}
\end{figure*}

The use of vision language models (VLMs) in education has received increasing attention in both academic research \cite{kuchemann2025opportunities, lee_interactive} and commercial AI products. Examples of the latter include Google Classroom with Gemini integration,\footnote{\url{https://blog.google/outreach-initiatives/education/classroom-ai-features/}} and Khan Academy's AI tutor Khanmigo,\footnote{\url{https://www.khanmigo.ai/}} powered by OpenAI models. However, the integration of these models into tutoring and classroom settings often lacks transparent, open, and realistic evaluation. With this gap in mind, we previously released DrawEduMath (Figure~\ref{fig:benchmark_example}), a dataset consisting of 2,030 teacher-annotated images of real students' hand-drawn responses to K-12 math problems \cite{baral-etal-2025-drawedumath}. In contrast to other multimodal math understanding or problem solving benchmarks \cite{alshammari2026mathnet}, DrawEduMath involves noisy, naturalistic data pulled from an online learning platform (Figure~\ref{fig:benchmark_example}). In the year since its release, we continuously updated the benchmark's leaderboard\footnote{\url{https://drawedumath.org/}} with newer models. 

\begin{figure*}[t]
    \includegraphics[width=\textwidth]{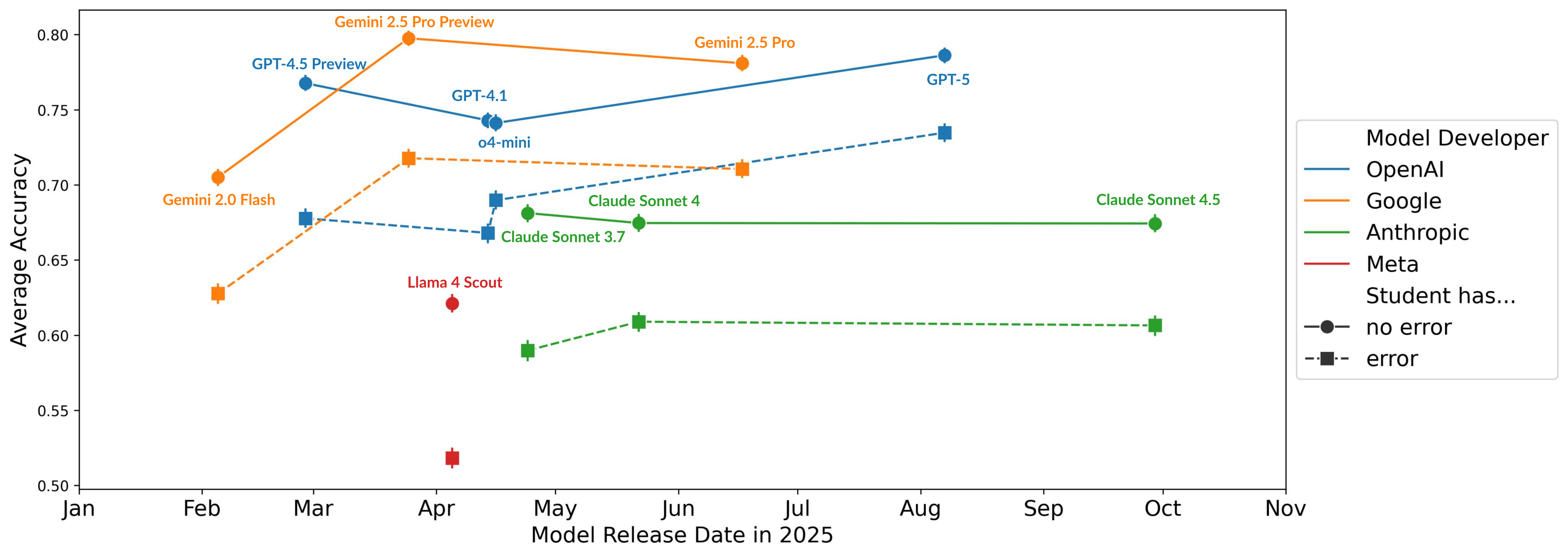}
    \centering
    \caption{VLMs consistently perform worse on answering DrawEduMath benchmark questions pertaining to erroneous student responses. Performance on non-erroneous student responses ($\iscircle$) is labeled with specific VLMs' names; that same model's performance on erroneous student responses is directly below ($\square$). Error bars are 95\% CI.}
	\label{fig:fig_1}
\end{figure*}

Our paper offers a snapshot of how 11 VLMs have performed on DrawEduMath in the year after its release (Figure~\ref{fig:fig_1}). We surface two key findings:

\begin{enumerate}[itemsep=0pt]
    \item[\textbf{F1}:] VLMs are worse at describing the contents of student work that contains math errors than student work without errors. 
    \item[\textbf{F2}:] VLMs still struggle the most on question types related to assessing students' correctness. 
\end{enumerate}

These findings suggest that VLMs underperform with students who need additional pedagogical support (\textbf{F1}), and they also fail to appropriately identify cases when support is needed (\textbf{F2}).

To investigate these patterns further, we conduct five analyses in \S\ref{sec:problem_effects}-\S\ref{sec:open_binary} targeting factors that may relate to model performance. Our experiments show that the performance gap in \textbf{F1} persists even when controlling for problem (\S\ref{sec:problem_effects}) and when image noise is reduced (\S\ref{sec:image_noise}). In addition, a possible explanation for \textbf{F1} is that VLMs expect mathematically correct input images. Indeed, we find that some models' wrongly predicted answers for erroneous student work are similar to gold answers for non-erroneous student work (\S\ref{sec:problem_defaults}). 

We also find that models do improve in assessing student correctness (\textbf{F2}) when provided gold natural language descriptions of student work (\S\ref{sec:infer_level}). However, their performance on these questions with extra textual support still lags behind their out-of-the-box performance on other question types. Finally, though models seemingly perform better on binary correctness questions (e.g. ``\textit{Does the student do \_\_\_ correctly?}'') than open-ended ones (e.g. ``\textit{What errors does the student make in their response?}''), some VLMs' performance can sometimes be barely better than chance (\S\ref{sec:open_binary}).

Altogether, our in-depth error analysis of VLMs' performance on real student math responses provides a clearer picture of their weaknesses in supporting K-12 math education.  We release data and scripts for reproducing our findings.\footnote{\url{https://github.com/lucy3/aftermath_drawedumath}}

\section{Background \& Related Work} \label{sec:related}

\paragraph{Multimodal math benchmarks.} Mathematical content, rich with diagrams, is ripe for evaluating models' multimodal abilities. Thus, many vision-language benchmark creation efforts have targeted math \cite{alshammari2026mathnet, zhang_mathverse, lu2024mathvista, yan-etal-2025-survey}. Educational settings offer additional challenges, where VLMs may be assessed on their abilities to make higher-level pedagogical inferences and handle handwritten, hand-drawn content \cite{parsaeifard2025automated, latif2025sketchmind, nath-etal-2025-vision, nguyen-etal-2025-vehme}. For example, MathCog asks models to diagnose students' cognitive skills using binary yes/no questions and a digitally handwritten dataset \cite{jin2025investigating}. Within this landscape of prior work, DrawEduMath remains a significant evaluation resource, given its diversity of image and question types, its use of noisy, real student work, and its inclusion of experienced teacher annotations \cite{baral-etal-2025-drawedumath}.

\paragraph{AI \& student error.} Student mistakes and misconceptions have long been a focal point in education research \cite{Smith_III01041994, radatz1979error, borasi1994capitalizing, metcalfe2017learning}. With increased attention towards AI as tutors and teaching assistants, research has focused on models' abilities to identify student error \cite{srivatsa-etal-2025-llms-spot, kochmar-etal-2025-findings, daheim-etal-2024-stepwise}, reason about patterned misconceptions \cite{rittle-johnson-etal-2025-detecting, ross2025learningmakemistakesmodeling}, correct errors \cite{mita-etal-2024-towards}, and provide feedback \cite{kaliisa2026does, botelho2023leveraging, stahl-etal-2024-exploring}. There is some research around model robustness to user error, but much of it focuses on linguistic errors in prompts \cite{gan-etal-2024-reasoning, chatterjee-etal-2024-posix}. There is little work on how math errors impact model performance: one example is \citet{daheim-etal-2024-stepwise}, who show that language models are worse at verifying the correctness of erroneous student math than non-erroneous math. Our work reaches a similar conclusion across more QA types and with multimodal data, though our results around models' correctness \& error assessments paint a less straightforward picture (\S\ref{sec:open_binary}). 

\paragraph{Risks of AI in education.} The field of AI \& education could be considered a form of AI for ``social good'' \cite{cowls2021definition}, driven by goals that counter AI's negative impacts on human skill formation and cognitive thinking \cite{bastani2025genAI, klimova2025exploring, shen2026ai, lee2025impact}. However, though its end-goals may be optimistically framed, AI in education is not without risk \cite{blodgett2021risks, holstein2021equity}. Our work is aligned with literature that investigates how AI may disparately impact different student populations \cite{schaller-etal-2024-fairness, capraro2024impact, educsci15050637}; our distinct approach is that we group student inputs based on demonstrated math proficiency. 

\section{Evaluating VLMs with DrawEduMath}

\subsection{Data} \label{sec:data}

DrawEduMath is an English-language dataset of 2,030 images of students’ handwritten responses to K-12 math problems \cite{baral-etal-2025-drawedumath}. Images are provided by the online learning platform ASSISTments and contain math problems drawn from open educational resources \cite{heffernan2014assistments, feng2025empowering}. Each image includes a math problem on the left and a student's response on the right, and is accompanied by three types of data: 

\begin{enumerate}[itemsep=0pt]
    \item \textbf{Free-form captions} (2.0k+) from teachers describing each student response image.
    \item \textbf{Synthetic QA pairs} (44.4k+), produced by Claude-3.5 Sonnet and GPT-4o reformatting facets of teachers' captions into QA, e.g. \textit{On the left-hand side of the image, the student wrote the word syrup} $\rightarrow$ \textit{What word did the student write on the left-hand side of the image? Syrup.}
    \item \textbf{Teacher-written QA pairs} (11.6k+). Teachers wrote a set of shared questions for each math problem, followed by answers for each student response to each problem. Teachers answered two additional generic questions, \textit{What errors does the student make in their response?} and \textit{What strategy does the student use to solve the problem?} across all problems and student responses. 
\end{enumerate}

DrawEduMath includes a taxonomy of seven question types. In our analysis, we simplify this taxonomy into three types: \textit{image creation and medium} (12.3\%), \textit{correctness \& errors} (8.5\%), and \textit{content description} (79.2\%). The first two match the benchmark's original taxonomy, while the third question type is an aggregation of all other question types, ranging from the student's problem solving strategy to the meaning, positioning, and frequency of drawn/written elements. We aggregate these question types in our main text, because our main findings generalize across more fine-grained types (Appendix~\ref{sec:disagg_results}). The original DrawEduMath paper includes additional dataset statistics and QA examples \cite{baral-etal-2025-drawedumath}. 

\subsection{Evaluation Setup} \label{sec:eval_setup}

\paragraph{Scoring metric.} \citet{baral-etal-2025-drawedumath} used Mixtral 8x22B to judge the similarity of VLMs' generated answers to gold answers on a scale of 1 (\textit{quite different answers}) - 4 (\textit{basically the same}), and then binarized these ratings when computing models' accuracy. As LMs update over time, older ones like Mixtral 8x22B become deprecated in model API services. Thus, all evaluation in our work uses an updated LLM judge. We take the majority vote from three judges: Claude Sonnet 4.5, Gemini 2.5 Pro, and GPT-4o.\footnote{Though GPT-4o is an older model, it has higher individual correlation with human annotations than the two newer models. So, we included it in our set of judges.} Our updated judge achieves similar correlation (Spearman $\rho$ = 0.808) with the same set of human judgements as \citet{baral-etal-2025-drawedumath}'s original judge ($\rho$ = 0.801). We compute and report binarized model accuracies; 1-2 are counted as incorrect, and 3-4 are correct. 

\paragraph{Models.} We evaluate 11 VLMs released in 2025 on DrawEduMath. Models span four developers: Open AI (GPT-4.1, GPT-4.5 Preview, o4-mini, GPT-5), Anthropic (Claude Sonnet 3.7, Claude Sonnet 4, Claude Sonnet 4.5), Google (Gemini 2.0 Flash, Gemini 2.5 Pro, Gemini 2.5 Pro Preview,), and Meta AI (Llama 4 Scout). Though our main findings \textbf{F1} \& \textbf{F2} pertain to all of these models, the analyses in \S\ref{sec:image_noise}-\S\ref{sec:open_binary} focus on four representative models: Gemini 2.5 Pro, Claude Sonnet 4.5, GPT-5, and Llama 4 Scout.

\paragraph{Labeling student error.} To categorize whether students' math responses contain an error or not, we use teachers' answers to the question, \textit{What errors does the student make in their response?} We ask GPT-5-mini to interpret each open-ended answer and classify it as \textit{yes}, as in, the teacher describes some error, or \textit{no}, for when teachers' answers are variations of \textit{There is no error} or \textit{The student did not make an error} (Appendix~\ref{sec:ann_error}). We validate this LM annotator on a manually checked random sample of 200 examples (F1 = 0.984).

\subsection{Main Findings} \label{sec:main_results}

\begin{figure}[t]
    \includegraphics[width=\columnwidth]{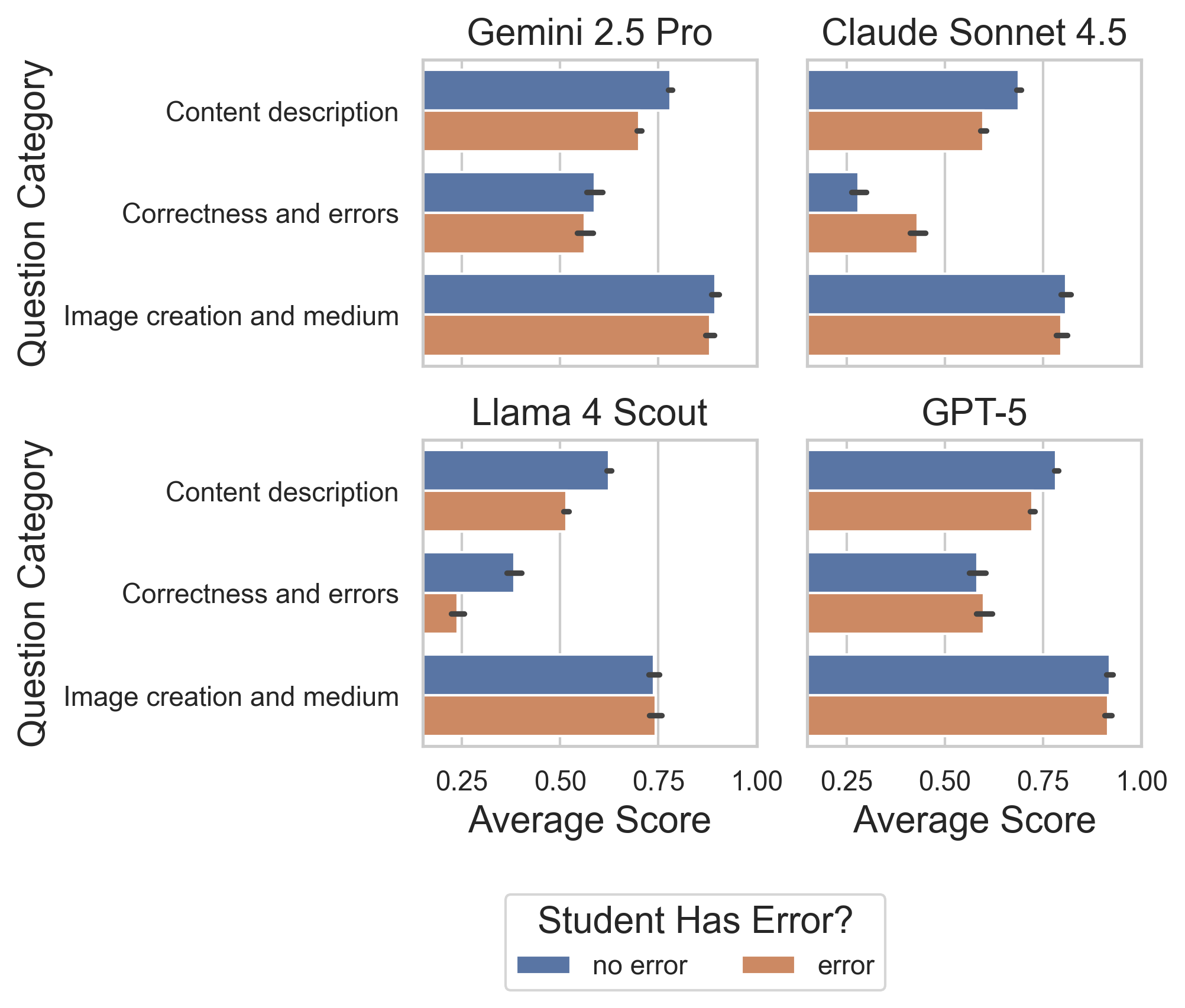}
    \centering
    \caption{Content description QA consistently drives the gap in VLM performance between student responses that contain errors versus those that do not. Appendix~\ref{sec:disagg_results} includes additional VLMs that expand this finding.}
	\label{fig:results_heatmap}
\end{figure}

On average, models tend to perform worse when the student response contains an error (Figure~\ref{fig:fig_1}). We find that this pattern is mostly driven by content description QA (\textbf{F1}, Figure~\ref{fig:results_heatmap}). In addition, a weakness reported by \citet{baral-etal-2025-drawedumath} in older VLMs persists in newer ones: questions related to students' correctness and errors are still the most difficult (\textbf{F2}). In the next few sections, we dive into five factors that we hypothesize to relate to these findings: problem effects (\S\ref{sec:problem_effects}), image noise (\S\ref{sec:image_noise}), problem response defaults (\S\ref{sec:problem_defaults}), visual understanding bottlenecks (\S\ref{sec:infer_level}), and question open-endedness (\S\ref{sec:open_binary}). 

\section{Models underperform on erroneous student responses even when controlling for problem} \label{sec:problem_effects}

\begin{table}[t]
\centering
\small 
\resizebox{0.9\columnwidth}{!}{%
\begin{tabular}{@{}p{4.5cm}p{2.5cm}@{}}
\toprule
\textbf{Model} & \textbf{$\beta_1$}\\ 
\midrule
Gemini 2.0 Flash &  0.0944*** \\ 
Gemini 2.5 Pro Preview & 0.0888*** \\
Gemini 2.5 Pro &  0.0889*** \\ \midrule
GPT-4.1  &  0.0865*** \\ 
GPT-4.5 preview & 0.0874***\\
o4-mini & 0.0612*** \\ 
GPT-5 & 0.0585*** \\ \midrule
Claude Sonnet 3.7  & 0.0922*** \\ 
Claude Sonnet 4  & 0.0816*** \\ 
Claude Sonnet 4.5 & 0.0842*** \\ \midrule
Llama 4 Scout & 0.1013*** \\ \bottomrule
\end{tabular}%
}
\caption{Estimated effects of student correctness ($\beta_1$) on VLMs' accuracy on content description QA, where all $p < 1.0^{-12}$.}
\label{tab:regression}
\end{table}

One possibility is that the model performance gap observed by \textbf{F1} is actually not affected by the presence or absence of student error, but rather by some math problems being more difficult for VLMs to understand. DrawEduMath contains 188 unique math problems targeting concepts ranging from geometry to fractions. These problems span multiple grade levels, and on average, each problem has 12.64 student responses \cite{baral-etal-2025-drawedumath}. Here, we show that the effect of student error on VLMs' content description QA performance is statistically significant even when controlling for problem. 

We estimate an ordinary least squares regression with problem fixed effects: 
$$y_{ij} = \beta_0 + \beta_1(s_{ij}) + u_j + \epsilon_{ij}$$
In the equation above, $y_{ij}$ is the average score a model has across content description QA for a student response $i$ and problem $j$, $u_j$ is a fixed effect for each problem, and $\epsilon_{ij}$ is the residual. If the student response is correct, $s_{ij}$ = 1, otherwise $s_{ij}$ = 0. Table~\ref{tab:regression} presents $\beta_1$ values across VLMs. These values show that even after controlling for problem, non-erroneous student responses significantly correspond with higher VLM performance. 

\section{Models' performance gaps are not strongly impacted by image noise} \label{sec:image_noise}

Another possible explanation for \textbf{F1} is that students who make math errors may simply submit noisier images. Students on ASSISTments may submit their answers by drawing digitally, or by uploading photographs of pen \& paper work, which may include smudges and blur. In this section, we ask, does the model performance gap described by \textbf{F1} remain even when students' responses are redrawn on a digital canvas in a standardized manner? 

\subsection{Experimental Setup}

Redrawing images is a time-intensive process, requiring careful interpretation of each math problem and the intent of the original student response. So, for this experiment, we stratify sample one erroneous student response and one correct response from each problem, yielding 336 images in total. Though this sample is small, it retains statistically significant gaps in VLM performance on content description QA between erroneous and non-erroneous student response images (Table~\ref{tab:noise_gap}). 

\begin{figure}[t]
    \includegraphics[width=0.75\columnwidth]{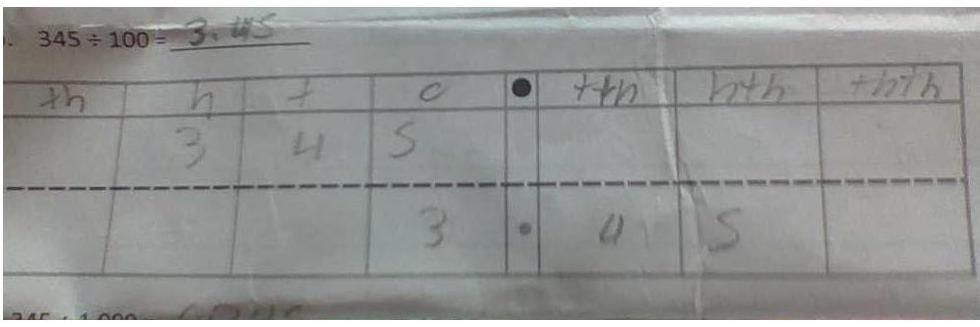}
    \includegraphics[width=0.75\columnwidth]{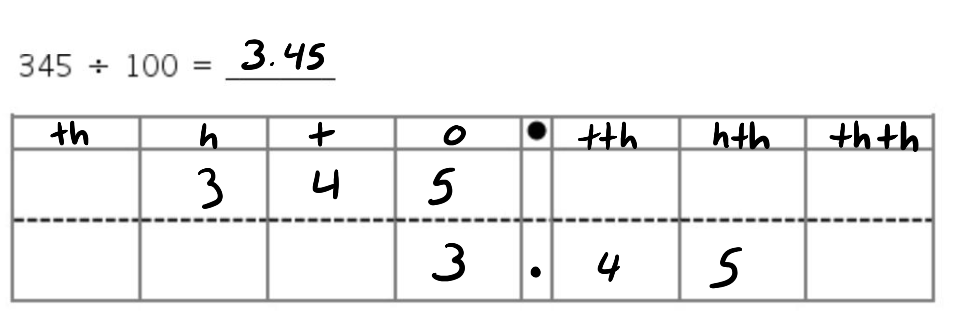}
    \centering
    \caption{An example of how a student response image (top) is transformed and cleaned up by our digital redrawing process (bottom). This student uses a place value chart to show how digit values change for 345 after division by 100.}
	\label{fig:redrawn}
\end{figure}

To encourage consistency, the lead author redrew all sampled student responses using a digital pen and the drawing application Procreate. This redrawing author retained students' original positioning of content, and consulted teachers' captions of images to navigate ambiguity and avoid faulty interpretation of problems and student responses. If needed, the author recreated elements such as graph paper grids and typed content in Figma. Figure~\ref{fig:redrawn} provides an illustrative example of how redrawing transforms students' responses.  

\subsection{Results}

\begin{figure}[t]
    \includegraphics[width=\columnwidth]{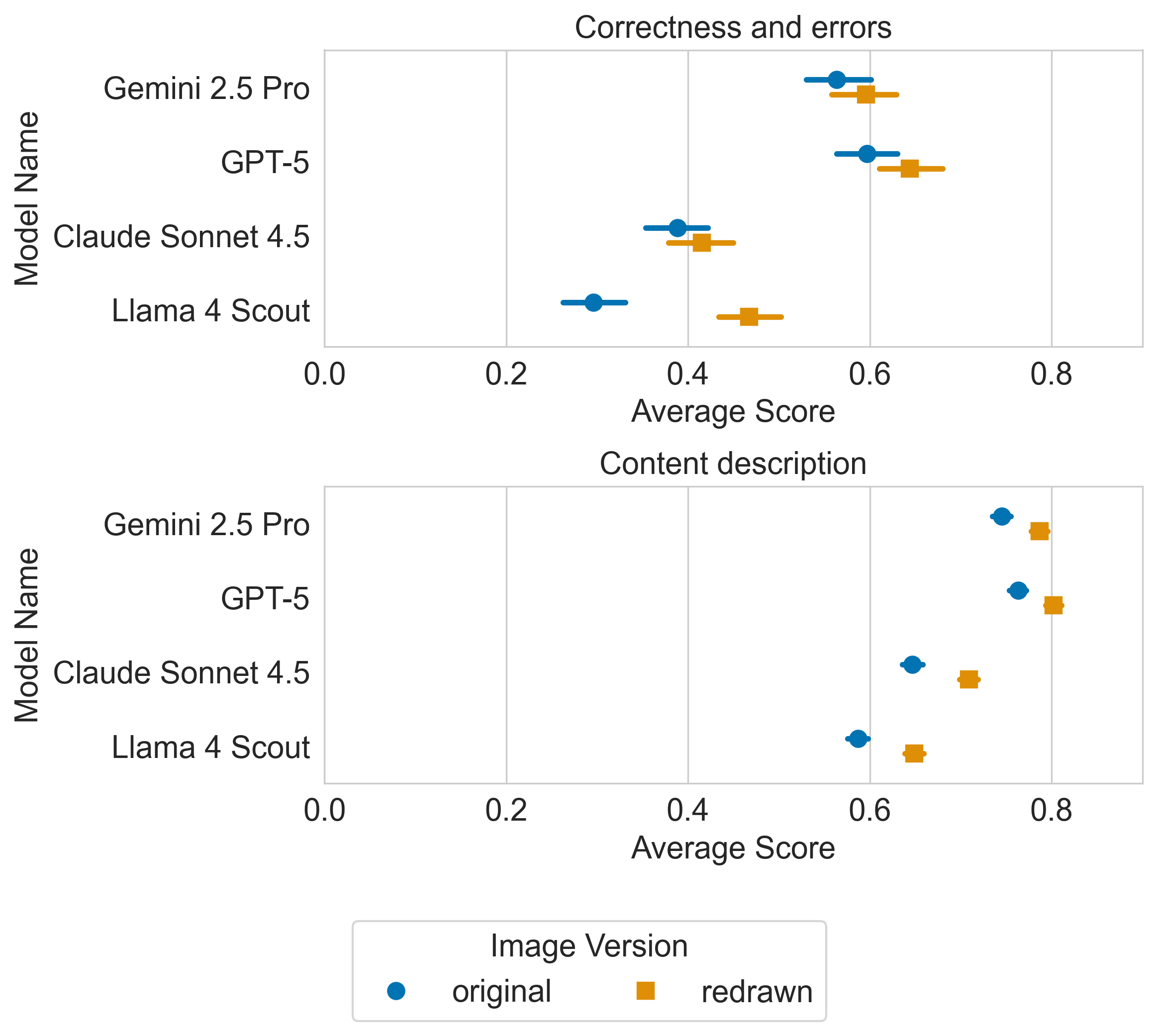}
    \centering
    \caption{Models' performance for content description QA generally improves after images are redrawn. Error bars are 95\% CI.}
	\label{fig:noise_improve}
\end{figure}

\begin{table}[t]
\centering
\resizebox{0.9\columnwidth}{!}{%
\begin{tabular}{@{}lcc@{}}
\toprule
\textbf{Model} & \textbf{Original Images} & \textbf{Redrawn Images}\\ 
\midrule
Gemini 2.5 Pro &  -0.089***  &  -0.096***  \\ 
Claude Sonnet 4.5 &  -0.100*** &  -0.058**  \\ 
GPT-5  & -0.055**  & -0.043**  \\ 
Llama 4 Scout &  -0.092***  &  -0.073*** \\ 
\bottomrule
\end{tabular}%
}
\caption{Differences in average scores on content description QA between erroneous and non-erroneous student images persist after redrawing. **$p<0.01$, ***$p<0.001$.}
\label{tab:noise_gap}
\end{table}

On redrawn images, VLMs' performance shift in the expected direction, where scores generally improve with less image noise (Figure~\ref{fig:noise_improve}). Though requesting students to only submit born-digital content may make their work more interpretable for AI, not all classrooms have resources and policies that make such standardization feasible. In addition, pen-and-paper work remains vital, with studies showing that this traditional mode of learning can sometimes allow students to surpass their digital-only peers \cite{mueller2014pen, altamura2025new, anthony2007benefits, umejima2021paper}. Thus, one implication of Figure~\ref{fig:noise_improve} is that the integration of VLMs in education may disparately impact analog and digital learners. 

Importantly, we also find that models' performance gap between erroneous and non-erroneous student images remains in redrawn images (Table~\ref{tab:noise_gap}). This result complements that of \S\ref{sec:problem_effects}, by further isolating student error as a weak point in the use of VLMs in educational settings. Thus, mitigation efforts around \textbf{F1} should focus on improving models’ understanding of erroneous mathematical content, across all levels of image noise and medium types. 


\section{Models default to assuming error-free math solutions} \label{sec:problem_defaults}

Why might erroneous student images be so challenging for VLMs? During a manual examination of VLMs' QA errors, 
we observed that models sometimes produce plausible, though wrong, answers to benchmark questions, especially considering the context of the provided math problem (Figure~\ref{fig:default_example}). This observation suggests a possible explanation for \textbf{F1}: VLMs perform better on content description QA for error-free student responses, because models default to assuming error-free math solutions. 

\subsection{Analysis Setup}

To quantify our observation using existing DrawEduMath annotations, we filter for content description QA shared across different student response images for the same math problem. Then, for each incorrect model answer to questions pertaining to erroneous student responses, we compare the model's answer against true answers for non-erroneous student responses, and see whether the model's incorrect answer matches the majority of these true ones. We compare model answers using the ensemble LM judge from \S\ref{sec:eval_setup}. We only consider cases where we have at least two non-erroneous student response images associated with the given question, to ensure that we have sufficient signal of correct student behavior. 

\subsection{Results}

\begin{figure}[t]
    \includegraphics[width=\columnwidth]{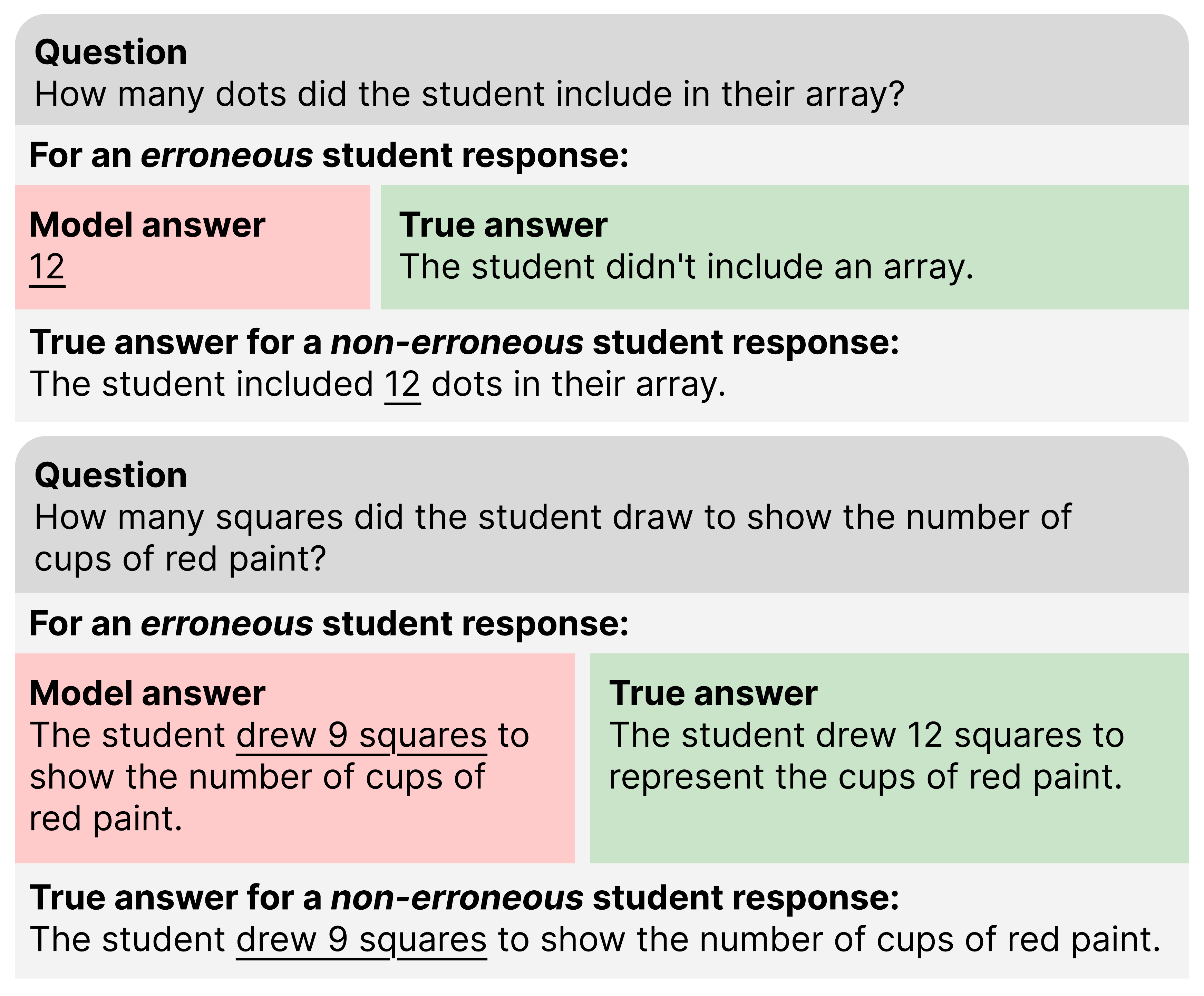}
    \centering
    \caption{Illustrative examples of the phenomenon where models predict answers for erroneous student responses that match true answers for non-erroneous students.}
	\label{fig:default_example}
\end{figure}

Across four representative VLMs, incorrect model responses for erroneous student responses sometimes do match non-erroneous student solutions, with percentages ranging from 29\% of content description QA mistakes for Gemini 2.5 Pro to 35\% for Claude Sonnet 4.5 (Table~\ref{tab:defaults}). So, a sizable portion of benchmark answers may be inferrable based on a math problem and a typical correct solution. There are many more ways a student response can be wrong than it can be correct, and so benchmark QA corresponding to correct student solutions navigate a narrower space of plausible possibilities. 

\begin{table}[t]
\centering
\small 
\resizebox{0.9\columnwidth}{!}{%
\begin{tabular}{@{}p{4.5cm}p{2.5cm}@{}}
\toprule
\textbf{Model} & \textbf{\%}\\ 
\midrule
Gemini 2.5 Pro &  0.2923 \\ 
Claude Sonnet 4.5 &  0.3519 \\ 
GPT-5  &  0.3125 \\ 
Llama 4 Scout &  0.3060 \\ \bottomrule
\end{tabular}%
}
\caption{The percentage of times for which an incorrect model answer for a content description question and erroneous student image matched the majority (> 50\%) of true answers for non-erroneous student images.
}
\label{tab:defaults}
\end{table}

Qualitatively, we observe that models especially tend to predict incorrect answers that match correct problem solutions when benchmark questions involve false presuppositions. Figure~\ref{fig:default_example} illustrates an example; there, the wording of the top teacher-written question assumes that the student has drawn any array at all. Teacher-written questions in DrawEduMath are those that teachers would like VLMs to answer across all student responses to a problem, mimicking potential uses of VLMs for learning analytics. Models' susceptibility to false premises or suppositions is well-documented in prior work \citep[e.g.][]{yu-etal-2023-crepe, srikanth-etal-2024-pregnant}, and our work illustrates a consequence of this weakness for education-related applications.

Generally, language models are developed to be good math problem solvers. Math solving benchmarks are continuously emphasized in leaderboards and commercial LM releases \cite{cobbe2021training, hendrycks2021measuring, gemini3pro2025}. To encourage mathematically correct outputs and hill-climb on these benchmarks, models are mostly exposed to ``high quality'', correct math content during training \cite{mahabadi2025nemotron, paster2024openwebmath}. The challenge of understanding, but not generating faulty content has received attention in other domains. For example, toxicity is another case of an understanding vs. generation tradeoff; we want models that can detect, address, and understand toxic content, without generating it \cite{longpre-etal-2024-pretrainers, wang2025teaching}. Our findings suggest that education is another domain where the application of alternative training methods on erroneous data, such as \citet{wang2025teaching}, could be applicable. 

\section{Textual support can improve models' correctness assessments to some extent} \label{sec:infer_level}

DrawEduMath QA range from low-level content description (e.g. ``\textit{How many triangles did the student draw?}'') to higher-level correctness judgements. Now, we move on from examining \textbf{F1}, which focuses on content description QA, to digging deeper into \textbf{F2}, which pertains to correctness \& errors QA. Earlier, we saw that the latter remain difficult even after images are digitally cleaned up (Figure~\ref{fig:noise_improve}). Perhaps, image understanding is a bottleneck for models answering these more reasoning-intensive questions. To what extent can models improve their assessment of student error when given textual descriptions of student work? 

\subsection{Experimental Setup} \label{sec:infer_level_setup}

As mentioned in \S\ref{sec:data}, DrawEduMath includes teacher-written, gold captions of students' response images. \citet{baral-etal-2025-drawedumath} used these captions to synthetically generate a subset of the QA pairs in the benchmark. If a caption produces synthetic QA that fall in the correctness \& error question category, we exclude that image from the current analysis to avoid input-output contamination.\footnote{We remove 262 images out of a total of 2,030. We considered editing captions rather than remove images, but correctness-related content was sometimes integrated with other caption content and would require intensive rewriting.} We append to each DrawEduMath input prompt these gold captions (prompt in Appendix~\ref{sec:nld_prompt}), and re-evaluate models' performance on correctness \& error QA. In addition, we also evaluate a setup where we ask models to generate their own descriptions of students' responses, and provide those captions in place of gold ones. 

\subsection{Results}

\begin{figure}[t]
    \includegraphics[width=\columnwidth]{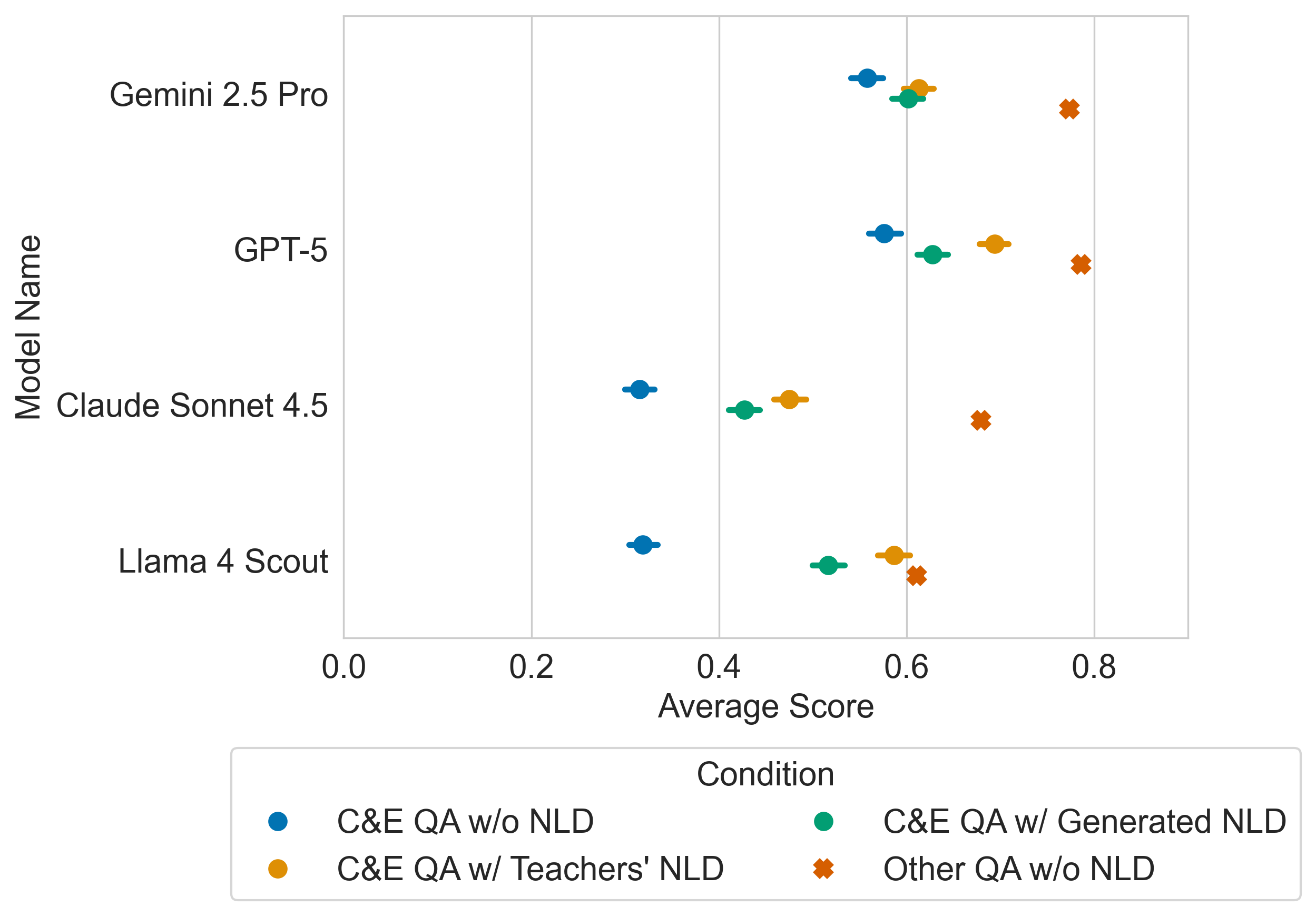}
    \centering
    \caption{Model performance on correctness \& error (C\&E) QA, with and without natural language description (NLD) support. We evaluate with a subset of input images and captions as described in \S\ref{sec:infer_level_setup}. Error bars are 95\% CI.}
	\label{fig:nld_results}
\end{figure}


Results are in the expected direction, in that VLM performance on correctness \& errors QA improves with natural language support (Figure~\ref{fig:nld_results}). However, this improved performance on correctness \& errors QA with captions still lags behind VLMs' caption-less performance in all other question categories for the same set of images. So, it is challenging for VLMs to make higher-level inferences useful for pedagogy even with gold textual support. A fully automatic two step caption-then-answer-QA process is a form of test-time scaling. We find that providing models their own generated captions of images can get models close to, but not match, their performance with teacher-written gold captions (Figure~\ref{fig:nld_results}). 

\section{Binary judgements of student correctness remain challenging} \label{sec:open_binary}


Open-ended questions are inherently more difficult than binary (e.g. yes/no) questions, as the latter is more guessable. Correctness \& errors QA (\textbf{F2}) in DrawEduMath provide a natural testbed for comparing open-ended questions (``\textit{What errors does the student make in their response?}'') and binary questions that assess whether some aspect of a student's response is correct/incorrect. 

\subsection{Analysis Setup} \label{sec:disagg_c_e_qa}

Our analysis splits correctness \& error QA into the following three subcategories: 

\begin{itemize}[itemsep=0pt]
    \item \texttt{Generic} questions (45.0\%). This is the open-ended question that DrawEduMath includes for all student images: ``\textit{What errors does the student make in their response?}''
    \item \texttt{Binary} assessments of specific solution components (50.4\%), e.g. ``\textit{Does the student put the decimal in the correct place in the product?}''
    \item \texttt{Other} questions (4.5\%), which mostly pertain to the nature of a student's error, e.g. ``\textit{What incorrect product did the student calculate for 667 times 5?}''
\end{itemize}

We use GPT-5-mini as an annotator to label whether non-\texttt{generic} questions are \texttt{binary} or \texttt{other} (prompt in Appendix~\ref{sec:ann_binary}). We validate this LM annotator on a manually labeled random sample of 200 unique questions (F1 = 0.975). 
We focus on \texttt{binary} and \texttt{generic} in the main text; Appendix~\ref{sec:add_correct_res} includes some results involving \texttt{other}.

Do models tend to predict that students make errors when they don't, or do they tend to overlook errors instead? For \texttt{binary} QA, we use GPT-5-mini to annotate whether questions and gold answers indicate that student is correct or incorrect (prompt in Appendix~\ref{sec:ann_student_correct}). For example, for the ground truth binary QA pair ``\textit{Q: Is there an error in the way that the number line has been drawn? A: Yes}'', the LM annotator would output that the student is \textit{incorrect}. We validate this LM annotator on a sample of 200 manually annotated random examples (F1 = 0.925). In total, the LM annotator labels 59.01\% of 2,274 binary QA as cases where the specified aspect of the student's response is correct, and the rest as ones where the specified aspect is incorrect. 

We also examine model performance on \texttt{generic} QA, disaggregated by whether the student's response is overall incorrect or correct. In DrawEduMath, successfully answering this question for correct students requires simply stating that there is no error, while for models to score well for erroneous students, they must also faithfully describe error specifics. 

\subsection{Results} \label{sec:binary_qa}

\begin{figure}[t]
    \includegraphics[width=\columnwidth]{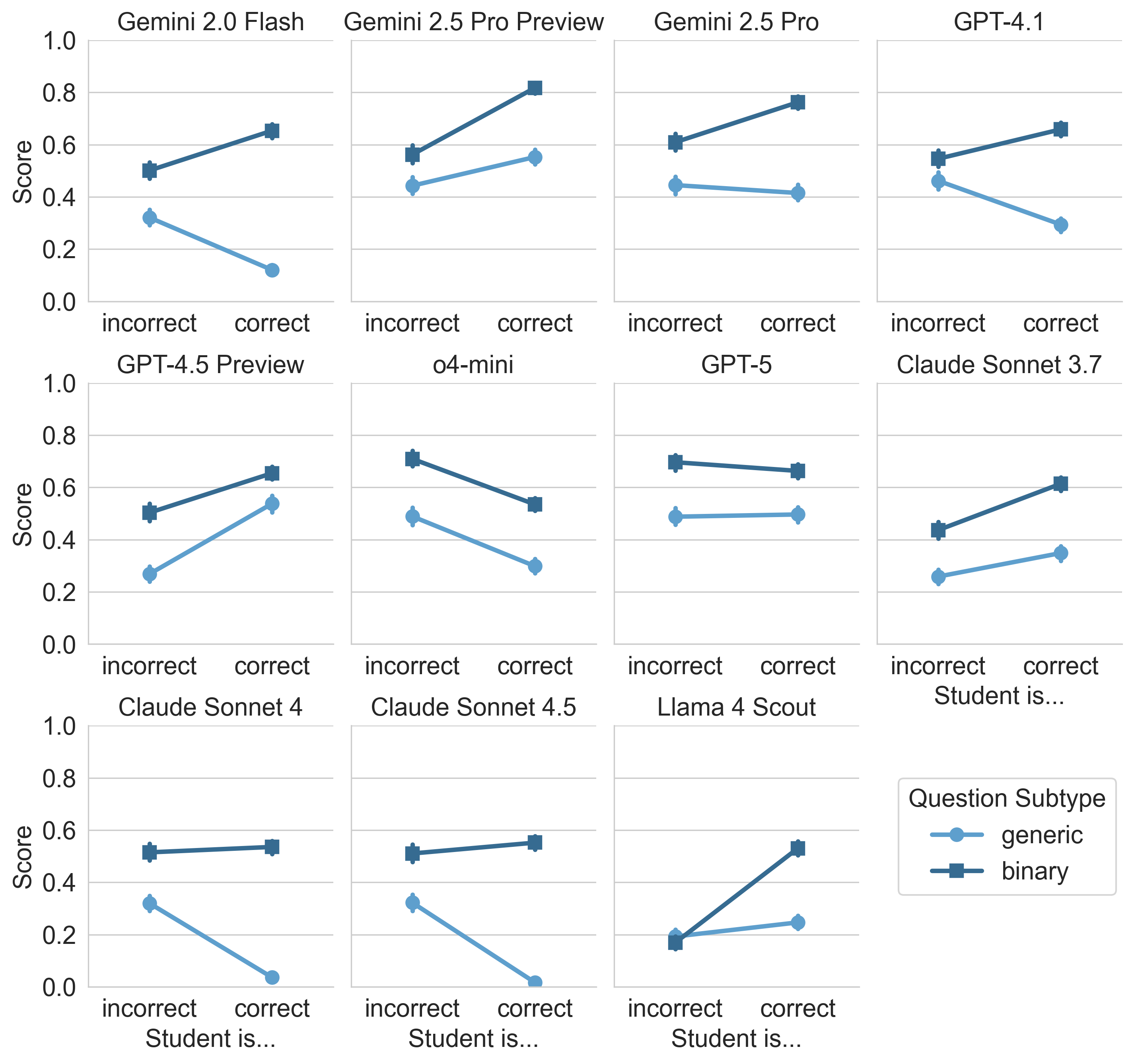}
    \centering
    \caption{VLMs' performance on the two main subtypes of correctness \& error QA, disaggregated by whether a student response is overall correct (\texttt{generic}) or correct based on specific aspect of their solution (\texttt{binary}). Error bars are 95\% CI.}
	\label{fig:binary_error_qa}
\end{figure}

Figure~\ref{fig:binary_error_qa} shows that performance patterns on correctness \& error QA, in relation to student correctness, generally and specifically, tend to be idiosyncratic. Some models tend to overreport errors being present, with lower scores on student images with \textit{no error}. Others struggle to detect and, in the case of \texttt{generic}, articulate errors that are present, with lower scores on student images with \textit{error}. Model behavior patterns are not shared across model versions from the same family or developer. Figure~\ref{fig:binary_error_qa} also indicates that some VLMs' \texttt{binary} QA scores hover closely around a random baseline of 0.5. Overall, assessing student error is incredibly challenging for VLMs, even though a substantial proportion of correctness \& error QA in DrawEduMath have high by-chance floor for performance.

\section{Conclusion}

Despite increasing attention towards the use of multimodal AI in education, our evaluation of 11 models released in 2025 demonstrates that their application on real student data remains challenging. Our findings suggest that erroneous student work is inherently more difficult for VLMs than correct student work (\textbf{F1}). VLM training and evaluation pipelines that favor correct mathematical content are at tension with the promise of AI for education, where incorrect math requires extra emphasis and attention. We also show that QA involving assessments of student correctness are particularly tricky (\textbf{F2}), across both text and image inputs, and across open-ended and binary question forms. 

Though this present paper presents a detailed error analysis of VLMs' performance on one vision-language K-12 math benchmark, our evaluation approach can be re-applied to other education-related benchmarks as well. That is, the evaluation of AI in education should be disaggregated in a manner that pinpoints whether models can actually discern when a student may need pedagogical support (\textbf{F2}), and whether they equitably serve students across different levels of proficiency (\textbf{F1}). Without a careful eye on the latter, models' capabilities may be overstated, and rushed integration into classrooms may exacerbate existing academic achievement gaps.

\section*{Limitations}

\paragraph{Scope and data representativeness.} Our study focuses on a single English benchmark, which involves student response images drawn from one online learning platform, ASSISTments. Thus, our findings may not map directly onto other languages and learning contexts. Based on school-level data provided by ASSISTments, we estimate that 85\% of images come from Title I schools, which are public schools in the U.S. that receive federal funding to support low-income students. ASSISTments partners with teachers and schools located across multiple location types (e.g. rural, suburban, town, city) and regions (e.g. West Coast, Midwest, East Coast, South), but self-selection is at play when it comes to which teachers, schools, and districts use the platform. DrawEduMath also contains questions that represent what was salient to Teaching Lab's teacher annotators \cite{baral-etal-2025-drawedumath}; it is not comprehensive of all of the ways in which educators may interpret and support student learning. Still, our high-level evaluation approach can be re-applied to other benchmarks and contexts, because transparency around the impact of student error on model performance is relevant to nearly all education-related settings. 

\paragraph{Data constraints.} Some of our experiments and analyses navigate practical, data-related constraints. For example, our image redrawing experiment in \S\ref{sec:image_noise} uses a small sample rather than the full dataset, since redrawing is a time-intensive process. Our other analyses rely on pre-existing teacher annotations and data present in DrawEdumath. For example, in \S\ref{sec:infer_level}, we removed some images from our analysis because their captions contained correctness \& error information, because models' performance with textual support on those images would be inflated. In addition, the content description QA under consideration for the results shown in Table~\ref{tab:defaults} are only questions shared across multiple student images for a problem, for which we could gather sufficient signal for what correct student response behavior should be. So, our results in that section (\S\ref{sec:problem_defaults}) primarily serve to illustrate one possible explanation for models' performance, and is not comprehensive of all of DrawEduMath. 

\section*{Ethical Considerations}

Education is a high-stakes setting for VLM use and deployment. The intermixing of AI and education involves delegating pedagogy to automated systems, impacting vulnerable underage populations, with possible life-long downstream effects related to economic mobility. Though there is optimism around AI's ability to support education \cite{DEMSZKY2025105183}, there should also be caution that it does not exacerbate existing inequities or introduce new ones \cite{Winters02012020, harvey2024towards}. We acknowledge that our work focuses primarily on technical harms measurable from model outputs, and does not capture broader harms that may emerge via interaction of AI with students, teachers, and school systems \cite{harvey2025dont}. In addition, AI research often involves a deployment-first mentality, where deployment may occur before a system has been deemed functional or necessary \cite{raji2022fallacy}. Our work advocates for robust evaluation and auditing of AI prior to deployment \cite{raji2020closing}, and accountability behind claims around model functionality and its social benefits \cite{kou2025dead, wang2024against}. 

\section*{Acknowledgments}

We are grateful for valuable feedback from Douglas Jaffe, who encouraged us to dig further into the impact of student error on model performance. We are also grateful for data-related support from Sami Baral, Neil Heffernan, and Cristina Heffernan. Our work is funded by the Gates Foundation. 

\bibliography{custom}

\appendix

\section{Model Performance by Question Type} \label{sec:disagg_results}

\begin{figure*}[t]
    \includegraphics[width=\textwidth]{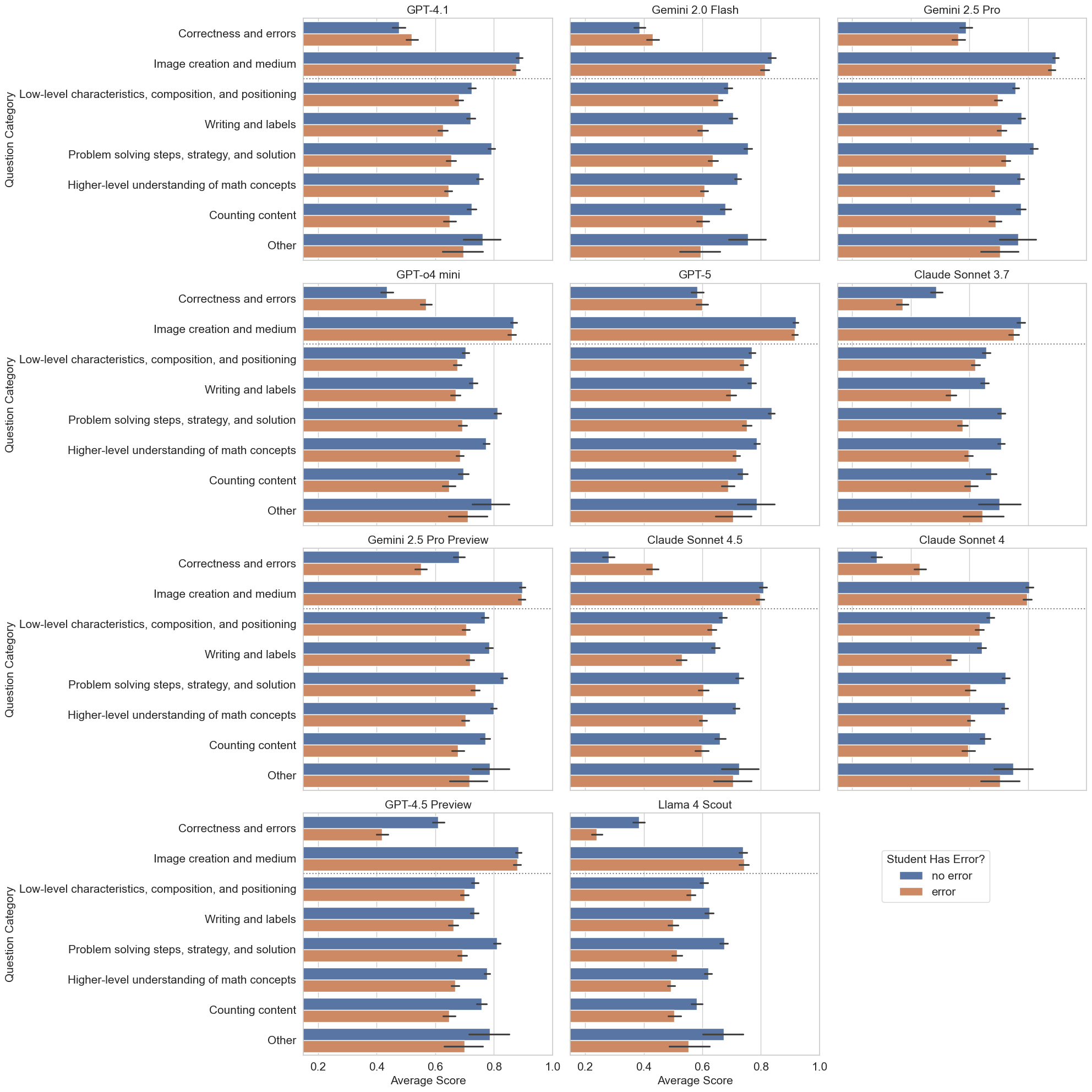}
    \centering
    \caption{An expanded version of Figure~\ref{fig:results_heatmap}, showing which question categories contribute to the gap in VLM performance between student responses that contain errors versus those that do not. Questions below the dotted line in each subplot are content description QA.}
	\label{fig:appdx_results_heatmap}
\end{figure*}

In the main text, Figure~\ref{fig:results_heatmap} shows that the gap in VLM performance between erroneous and non-erroneous student responses is primarily driven by content description QA. Figure~\ref{fig:appdx_results_heatmap} expands upon that finding, by disaggregating content description QA's overall pattern into finer-grained question categories and showing results for all 11 VLMs. Across these expanded plots, we see that the performance gap between erroneous and non-erroneous student responses persists across finer-grained content description QA categories. 

\section{Language Model-Assisted Data Annotation}

For each LM-assisted labeling task, we iteratively developed prompts that yield solid performance on small samples, before validating our final prompts on larger samples. The main text details the performance of each prompt on the intended task. 

\subsection{Student Error} \label{sec:ann_error}

One of our main findings, \textbf{F1}, pertains to how models perform differently between student responses that contain errors versus those that do not. To determine whether a student response contains an error or not, we rely on teachers' free-form descriptions of student error. Since teachers' written responses may span a variety of phrasings, we use GPT-5-mini to decisively label whether the teacher indicates that the student response contains an error. Here, \texttt{ans} is the teacher's answer to the question, \textit{What errors does the student make in their response?} We use the following prompt: 

\begin{lstlisting}
When asked about what errors a student makes in their response to a math problem, a teacher writes, '{ans}'. Based on the teacher's feedback, does the student make any error? Respond 'yes' or 'no'.
\end{lstlisting}

\subsection{Finer-grained Correctness \& Error Questions} \label{sec:ann_binary}

In \S\ref{sec:disagg_c_e_qa}, we discuss how correctness \& error QA span both binary assessments of specific student errors and more-open-ended questions. We use GPT-5-mini as an annotator to label whether each Correctness \& Error \texttt{question} is \texttt{binary} or \texttt{other}. We use the following prompt: 

\begin{lstlisting}
Is the following question a binary question that asks whether a student does something correctly or not?
Question: '{question}'
Decide whether the question above is a binary question that judges a student's correctness. Your response should start with 'Yes' or 'No':
\end{lstlisting}

\subsection{Binary Student Correctness} \label{sec:ann_student_correct}

In \S\ref{sec:binary_qa}, we're interested in investigating whether models tend to under- or over-report student error on binary correctness \& error questions. We use GPT-5-mini to annotate whether \texttt{binary} QA's questions and gold answers indicate that student is correct or incorrect. We use the following prompt: 

\begin{lstlisting}
Teacher A is examining a student's solution to a math problem. Teacher B asks Teacher A, '{question}'
Teacher A says, '{answer}'.
Does this exchange indicate that the student's solution has an error? Respond "yes" or "no":
\end{lstlisting}

\section{Natural Language Description Experiments}

In \S\ref{sec:infer_level}, we investigate whether the inclusion of textual descriptions of students' work can support VLMs' abilities to make higher-level inferences around students' correctness and errors. 

\subsection{Prompts} \label{sec:nld_prompt}

Our prompt that adds in natural language descriptions/captions is intuitive and simple:

\begin{lstlisting}
Description of image: 
{caption}

Answer the following question: {question}
\end{lstlisting}

To generate descriptions or captions using language models, we use the following prompt: \texttt{Describe the Student Response on the right side of the image in one paragraph.}



\section{Additional Correctness \& Error Results} \label{sec:add_correct_res}

\begin{figure}[t]
    \includegraphics[width=\columnwidth]{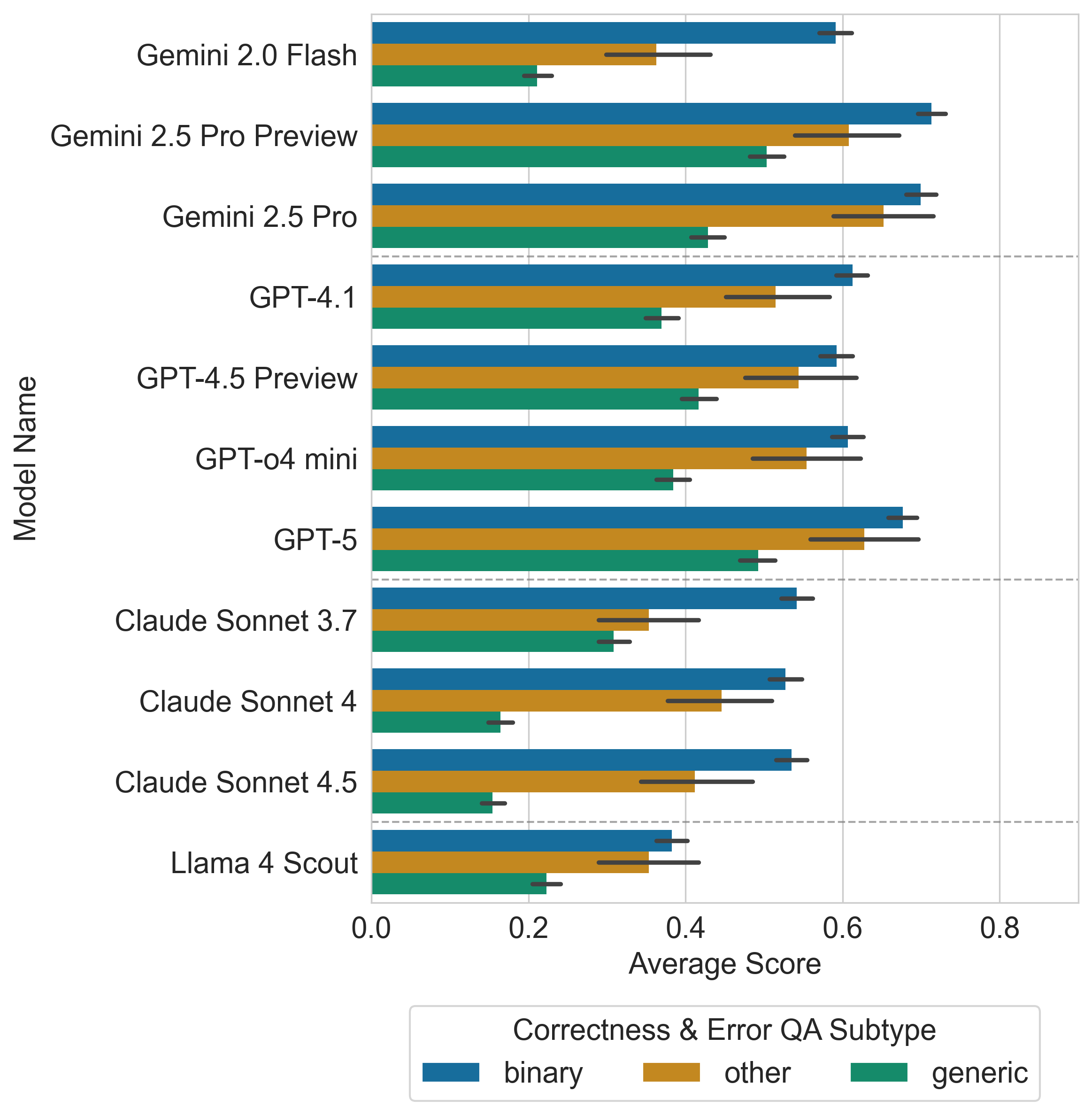}
    \centering
    \caption{VLMs' performance across different subtypes of correctness \& error questions, as defined in \S\ref{sec:disagg_c_e_qa}.
    }
	\label{fig:disagg_error_qa_all}
\end{figure}

The relative performance ranking of \texttt{binary}, \texttt{other}, and \texttt{generic} correctness \& error QA is consistent across all 11 VLMs (Figure~\ref{fig:disagg_error_qa_all}).

\end{document}